\documentclass{easychair}

\usepackage{graphicx}   % \includegraphics
\usepackage{caption}    % \captionof
\usepackage{doc}
\usepackage{titlesec}
\usepackage{enumitem}
\titlespacing*{\section}{0pt}{5pt plus 1pt minus 1pt}{5pt}

\usepackage[utf8]{inputenc}
\usepackage{CJKutf8}
\usepackage{tabularx}
\usepackage[bottom]{footmisc}
\begin{document}
\begin{CJK}{UTF8}{mj}

\title{Explainable Adversarial-Robust Vision-Language-Action Model for Robotic Manipulation%
\thanks{This research was supported by the Regional Innovation System \& Education(RISE) program
through the RISE Center, Gyeongsangnam-do, funded by the Ministry of Education(MOE) and
the Gyeongsangnam-do Provincial Government, Republic of Korea. (2025-RISE-16-001)}
 \vspace{-4pt}}

% Authors are joined by \and. Their affiliations are given by \inst, which indexes
% into the list defined using \institute
%

\author{
   \vspace{-4pt}
    Ju-Young Kim\inst{1}
\and
    Ji-Hong Park\inst{1}
\and
    Myeongjun Kim\inst{2}
\and
    Gun-Woo Kim\inst{1}\thanks{Corresponding author}
}
% Institutes for affiliations are also joined by \and,
\institute{
  Department of Computer Science and Engineering
\and
  Department of AI Convergence Engineering\\
  Gyeongsang National University, Jinju, Republic of Korea \\
  \email{\{wndudwkd003, hong\_0002, gnu\_kim98, gunwoo.kim\}@gnu.ac.kr}
}

%  \authorrunning{} has to be set for the shorter version of the authors' names;
% otherwise a warning will be rendered in the running heads. When processed by
% EasyChair, this command is mandatory: a document without \authorrunning
% will be rejected by EasyChair

\authorrunning{J.-Y. Kim et al.}
\titlerunning{Explainable Adversarial-Robust VLA Model for Robotic Manipulation}

\maketitle

\vspace{-12pt}
\setlength{\skip\footins}{8pt}
\begin{abstract}
Smart farming has emerged as a key technology for advancing modern agriculture through automation and intelligent control.
However, systems relying on RGB cameras for perception and robotic manipulators for control, common in smart farming, are vulnerable to photometric perturbations such as hue, illumination, and noise changes, which can cause malfunction under adversarial attacks.
To address this issue, we propose an explainable adversarial-robust Vision-Language-Action model based on the OpenVLA-OFT framework.
The model integrates an Evidence-3 module that detects photometric perturbations and generates natural language explanations of their causes and effects.
Experiments show that the proposed model reduces Current Action L1 loss by 21.7\% and Next Actions L1 loss by 18.4\% compared to the baseline, demonstrating improved action prediction accuracy and explainability under adversarial conditions.

\vspace{8pt}
\end{abstract}

\begin{center}
   \vspace{-8pt}
    \includegraphics[width=0.95\textwidth]{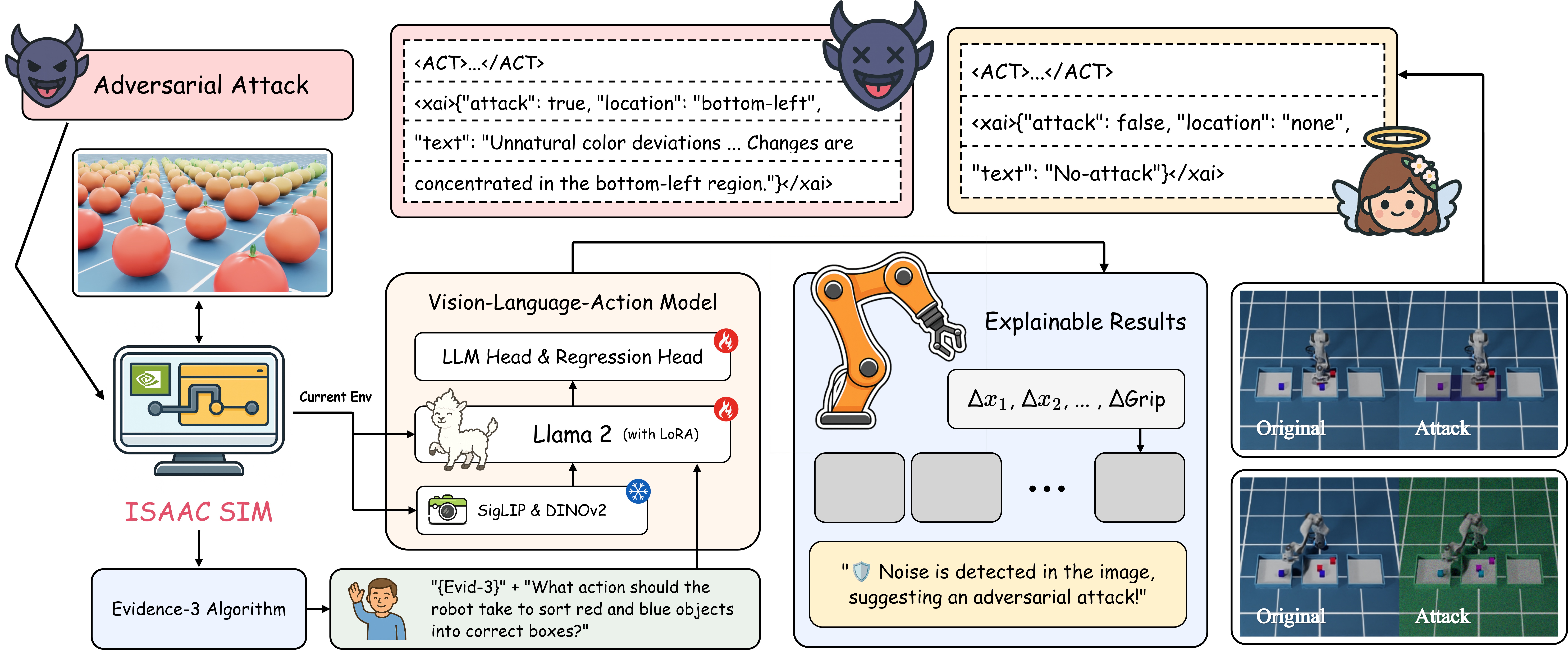}
    \captionof{figure}{Overview of the proposed architecture.}
    \label{fig:overview}
\end{center}
\vspace{-12pt}
\section{Introduction and Proposed Method}
  \vspace{-2pt}
\label{sect:introduction}
Smart farming systems are complex intelligent systems that integrate various modules such as robots, sensors, and cameras.
While Vision-Language-Action (VLA) research has advanced in processing multimodal data for environmental perception and control,
studies on explainable artificial intelligence (XAI) for defending against adversarial attacks remain limited\cite{GAO2024109557}.
This paper proposes a VLA model capable of detecting and explaining adversarial attacks by integrating an adversarial detection and explanation module into the existing VLA framework, as illustrated in Figure~\ref{fig:overview}.
The following section outlines the proposed architecture.
\vspace{5pt}
\begin{enumerate}[topsep=3pt, itemsep=2pt, parsep=2pt]
    \item \textbf{Adversarial Data Generation:}
    Simulation data are collected using the Franka Emika Panda robotic arm and an RGB camera in Isaac Sim.
    Random photometric transformations including hue shift ($T_{\mathrm{color}}$), illumination adjustment ($T_{\mathrm{illum}}$), and noise injection ($T_{\mathrm{noise}}$) are applied to generate adversarial variants.
    Formally, this process can be expressed as
    $x' = T_{S}(x), \; T_{S}\subseteq\{T_{\mathrm{color}},\,T_{\mathrm{illum}},\,T_{\mathrm{noise}}\}$,
    where $x$ denotes the original input image, and $S$ is a randomly selected subset of transformations.
    \item \textbf{Evidence-3 Module Integration:}
    The proposed architecture builds upon the OpenVLA-OFT framework~\cite{kim2025finetuningvisionlanguageactionmodelsoptimizing} and incorporates an additional Evidence-3 module for adversarial attack detection.
    The Evidence-3 module consists of a detection pipeline based on three statistical metrics: HSV Mahalanobis Distance (detecting color distribution anomalies), High-Frequency Energy Ratio (identifying noise injection), and Local Entropy Standard Deviation (capturing spatial irregularities).
    These statistical cues are embedded into the user instruction and provided as auxiliary input to the model.
    \item \textbf{Action Prediction and Explainable Model Training:}
    The action prediction head receives hidden representations from the Llama2 backbone and predicts the current and subsequent actions by minimizing the L1 loss.
    In parallel, the model is trained to detect and describe adversarial attacks by minimizing the cross-entropy loss over the XAI tokens generated by the Llama2 output.
    The total loss is defined as $\mathcal{L}_{\mathrm{total}} = \lambda_{\mathrm{xai}}\mathcal{L}_{\mathrm{xai}} + \mathcal{L}_{\mathrm{act}}$,
    where $\mathcal{L}_{\mathrm{xai}}$ represents the cross-entropy loss for explanation tokens,
    scaled by a weighting hyperparameter $\lambda_{\mathrm{xai}}$ that controls the relative importance of explanation learning, set to 0.5 in this work.
    Meanwhile, $\mathcal{L}_{\mathrm{act}}$ denotes the L1 regression loss for action prediction.
\end{enumerate}
  \vspace{-2pt}
\section{Experimental Results and Conclusion}
  \vspace{-2pt}
\label{sect:experiments}
To evaluate the proposed architecture, we compared three configurations: the baseline (Default), an adversarially trained model (Augmentd), and the proposed model.
Table~\ref{tab:results} summarizes the results.
Compared with the Default model, the proposed model reduced the Current and Next Action L1 losses by 21.6\% and 18.4\%, respectively, while outperforming the Augmented model by 6.9\% and 7.8\%, respectively.
It also achieved an XAI token accuracy of 99.77\%, showing that joint learning of robustness and explainability improves action prediction under adversarial conditions.
Future work will explore the applicability of our approach to real-world smart farming environments and extend validation using robotic simulations.
  \vspace{0pt}
\begin{table}[h!]

\centering
\caption{Performance evaluation results between the proposed and baseline models.}
\vspace{-2pt}
\label{tab:results}
\small
\begin{tabularx}{0.95\textwidth}{
  l
  >{\hsize=1.0\hsize\centering\arraybackslash}X
  >{\hsize=1.0\hsize\centering\arraybackslash}X
  >{\hsize=1.0\hsize\centering\arraybackslash}X
}
\hline
\textbf{\scriptsize OpenVLA-OFT} &
\textbf{\scriptsize \mbox{XAI Token Accuracy (\%)}} &
\textbf{\scriptsize \mbox{Current Action L1 ($\downarrow$)}} &
\textbf{\scriptsize \mbox{Next Actions L1 ($\downarrow$)}} \\
\hline
Default & - & 0.0826 & 0.0788 \\
Augmented & - & 0.0695 & 0.0697 \\
Proposed & \textbf{99.77} & \textbf{0.0647} & \textbf{0.0643} \\
\hline
\end{tabularx}
\end{table}
  \vspace{-5pt}
\label{sect:bib}
\bibliographystyle{abbrv}

\bibliography{easychair}
\end{CJK}
\end{document}